\documentclass[letterpaper]{article}

\pdfoutput=1
\usepackage[final]{neurips_2021}

\usepackage[utf8]{inputenc} 
\usepackage[T1]{fontenc}    
\usepackage{hyperref}       
\usepackage{url}            
\usepackage{booktabs}       
\usepackage{amsfonts}       
\usepackage{nicefrac}       
\usepackage{microtype}      
\usepackage{xcolor}         
\usepackage{tabu}
\usepackage{graphicx}

\title{Data Augmentation for Intent Classification}

\author{%
  Derek Chen, Claire Yin\footnotemark \\
  ASAPP, New York, NY 10017\\
  \texttt{\{dchen, cyin\}@asapp.com}
}
\begin{document}

\maketitle

\renewcommand*{\thefootnote}{\fnsymbol{footnote}}
\footnotetext{Work done while interning at ASAPP.}

\begin{abstract}
Training accurate intent classifiers requires labeled data, which can be costly to obtain. Data augmentation methods may ameliorate this issue, but the quality of the generated data varies significantly across techniques.  We study the process of systematically producing pseudo-labeled data given a small seed set using a wide variety of data augmentation techniques, including mixing methods together.  We find that while certain methods dramatically improve qualitative and quantitative performance, other methods have minimal or even negative impact.  We also analyze key considerations when implementing data augmentation methods in production.
\end{abstract}

\section{Introduction}
The performance of machine learning models is highly dependent on quantity of the data used to train it, but annotated data is often costly and time-consuming to collect. 
Data augmentation has emerged as a possible solution, where there has been significant progress in recent years across numerous categories~\citep{andreas20good, niu2019auto, gao20paraphrase}. Surface form alteration augmentation methods change the surface level text to produce new forms~\citep{wei2019eda}. Latent perturbation maps text to a hidden state before mapping back to natural language text again~\citep{zhao18vae}.  Auxiliary datasets take advantage of external unlabeled data from a relevant domain to form new pseudo-labeled examples~\citep{chen2021gold}. Text generation uses large pre-trained models to create new examples~\citep{devlin2018bert}. 

While these methods are promising, there is little understanding on how they compare individually and categorically, especially given real life consideration such as difficulty of implementation, model maintenance, and inference speed.  Data augmentation for natural language systems are particularly challenging since changing even a single word can change the meaning of the text~\citep{ng20ssmba}.  Due to this uncertainty and the cost of setting up the augmentation methods, most practitioners default to manual data annotation.  This may be more dependable, but is certainly not as scalable. 

In this paper, we first measure the impact of data augmentation on model performance with both quantitative and qualitative metrics. Next, we want to know how the benefits of different augmentation methods vary as parameters change, such as which specific method(s) are applied or what domain they are applied towards. 
Third, we experiment with using different combinations of methods together to see if mixing helps and to what extent it helps (ie. does order of mixing matter). Finally, we aim to understand the trade-off between model performance and model complexity. 

Our experiments find that while no category of methods works consistently, there are specific augmentation techniques which provide reliable benefit across different domains and settings.  At the same time, we also discover that certain methods perform so poorly that adding the augmentations cause the model to perform worse than if it had no extra data at all.  Furthermore, we report that mixing different augmentation sources can show strong results depending on the seed data and sources being combined. Overall, our results show that data augmentation methods are sensitive to various parameters, but can indeed be useful for real life systems if applied carefully.
\section{Methods and Experiments}
We study the impact of data augmentation applied to dialogue intent classification. The intent classification task takes an utterance as input and asks a model to predict the correct class from a finite set of intents. 
Data augmentation works in reverse: given an intent and a small seed set of utterances, a model is learned to produce utterances as output. 
The overall goal is to 
generate enough quality data to improve downstream model performance compared to only using the seed data. 
Our intent classifier consists of RoBERTa-base 
followed by a 2-layer MLP for prediction~\citep{liu2019roberta}. 

We study this problem for two domains: airline and telecom, which contain 128 and 118 intents, respectively.  Each method starts with five seed utterances per intent 
from which the augmentation method will attempt to produce 10x that amount. 
For airlines, this is 128 * 50 = 6,400 possible augmentations, or equivalently 7,040 training examples when including the seed data. 
Data augmentation methods can be roughly divided into four separate categories.  We experiment with two techniques for each category, and further mix techniques to form four new combinations, resulting 12 total methods. 

\subsection{Surface Form Alteration}
One way of augmenting natural language data is to alter the surface form of the text.  (1a) \textit{Easy Data Augmentation} (EDA) produces new examples by randomly deleting, inserting or swapping the order of tokens within the seed utterance~\citep{wei2019eda}.
(1b) \textit{Synonym Replacement} 
first determines the part-of-speech (POS) for each utterance token, and then chooses tokens for replacement only when the POS tag is either a noun, verb, or adjective.  Finding synonyms in this manner avoids stop words and produces more grammatical augmentations.  We use Wordnet as our source~\citep{miller1995wordnet}. 

\subsection{Latent Perturbation}
New data can also be generated by mapping the raw utterance into a latent embedding space and back into an alternate surface form.  
(2a) \textit{Back-translation} encodes the original utterance in a separate language, and then translates back into English to produce a new example~\citep{junczys2018mariannmt}.  We apply this technique using French, Portuguese, Spanish, German and Russian, where the languages were selected based on initial experiments.
(2b) We also pretrain a model to perform \textit{Paraphrasing} using the PAWS, QQP and MRPC corpora~\citep{zhang2019paws, iyer2017qqp, dolan2005mrpc}.  Specifically, we use a BART model which takes the original utterance as the input sequence and produces a paraphrased utterance as the output sequence~\citep{lewis2020bart}.  

\subsection{Text Generation}
Text Generation methods augment data by filling in novel words or characters based on its learned understanding of natural lanugage patterns.  We use (3a) \textit{Text In-filling} which first masks out random tokens from the original utterance, and then uses BERT to fill in the blanks~\citep{devlin2018bert}.  We additionally use (3b) a separate BART model from 2b to produce utterances with typos.  While a model could \textit{Generate Typos} by simply inserting 
random characters, our method fine tunes a large transformer model ~\citep{vaswani2017transformer} using the Github Typo corpus~\citep{hagiwara2020typo} which contains typos pulled from real Github commits, producing noticeably more realistic errors.
	
\subsection{Auxilliary Dataset}
While methods from prior categories are self-contained, the Auxiliary Dataset methods require access to a separate dataset of utterances coming from the same distribution.
(4a) \textit{k-Nearest Neighbors} retrieves new training utterances from a pool of unlabeled conversations.  Specifically, we embed dialogues turns as a bag-of-words with GloVe~\citep{pennington2014glove} to form our candidate pool.  We then embed each incoming query in the same manner and use Euclidean distance to find the nearest neighbor.  In practice, we use the FAISS library to speed up retrieval~\citep{johnson2017faiss}. (4b) Finally, we fine-tune a  \textit{Language Model} (LM) to decode new text given an intent.  Concretely, a dataset containing labeled utterances are fed into GPT2 ~\citep{radford20gpt}, where each example starts with the intent tag and ends with the utterance text, with a separator token in between.  During inference time, we feed only the intent tokens followed by the separator token and ask the model to hallucinate the utterance portion of the text.

\subsection{Mixing Methods}
We also generate training data using a combination of the different augmentation methods. 
(5a) \textit{Top 4 Accuracy} uniformly samples examples from the top four individual techniques to form the training set, as measured by model accuracy.  
(5b)  Next, we consider a mix consisting of \textit{Category Best}, which includes the better of the two techniques within each category. 
These are EDA, Paraphrase, Typo and LM Decoding. 
(5c) \textit{Heuristic Selection} takes advantage of reviewing the qualitative outputs. Specifically, we found during initial experiments that Paraphrase and LM Decoding produce highly varied sentence structure, while Typo makes minor changes to the text.  As such, we designed an interactive method that first performs Paraphrase and LM Decoding to generate half of the augmented examples and then applies Typo Generation to the augmented data to produce the remaining half. (5d) \textit{Mix All} applies any one of the eight techniques at random when producing each augmentation.
\section{Results}
\begin{table}
\centering
\resizebox{\textwidth}{!}{
\begin{tabu}{l|ccccc}
\textbf{Method} &  \textbf{\# Augment} $\uparrow$ & \textbf{Diversity} $\uparrow$ & \textbf{Accuracy} 
        $\uparrow$ & \textbf{Time Spent} $\downarrow$ & \textbf{Accept Rate} $\uparrow$ \\
\tabucline[1.4pt]{-}
Baseline           &   ---  & ---  & 62.9 & 11.8 & ---    \\
\hline
EDA                &  6280 & 29.4 &  72.5 & 1.52 & 55.3\% \\
Synonym            &  5909 & 29.9 &  65.0 & 1.46 & 60.6\% \\
Paraphrase         &  6344 & 35.4 &  84.9 & 1.43 & 69.4\% \\
Translation        &  2437 & 28.0 &  75.8 & 1.56 & 50.6\% \\
Text In-filling    &  6231 & 26.7 &  63.3 & 1.10 & 62.8\% \\
Typo Generation    & \textbf{6378} & 28.7 & \textbf{87.5} & \textbf{1.08} & \textbf{81.1\%} \\
kNN Retrieval      & 3392 & \textbf{51.7} & 55.8 & 1.63 & 11.9\% \\
LM Decoding        &  6152 & 35.2 &  83.3 & 1.24 & 34.2\% \\
\hline
Top 4 Accuracy     & \textit{6400} & 35.9 & 71.3 & 1.04 & \textit{68.9\%} \\
Category Best      & \textit{6400} & 36.4 & 70.4 & 0.98 & 66.7\% \\
Heuristic Select   & \textit{6400} & \textit{38.3} & \textit{80.0} & 0.98 & 58.3\% \\
Mix All            & \textit{6400} & 37.3 & 67.5 & \textit{0.90} & 64.4\% \\
\end{tabu}}
\caption{Results for airline domain. Baseline is an intent classifier trained only with seed data.  Top single method results are bolded, and top mixed method results are in italics. Accuracy represents the downstream model accuracy on intent prediction.  Time Spent is written out in minutes.} \label{tab:airline}
\end{table}

\subsection{Baseline Results}

\paragraph{Automatic Metrics}
We evaluate the different augmentation methods on three automatic metrics: output quality, output magnitude and output diversity. We define output quality as the downstream classifier test accuracy.
We define output magnitude by number of augmentations produced. 
Given the limited seed set, all methods occasionally produce duplicate augmentations.  Once ten duplicates occur for a given intent,  
the algorithm terminates, so higher output magnitude serves as a proxy for the method's reliability. Lastly, we define output diversity through the Measure of Textual Lexical Diversity (MTLD)~\citep{mccarthy2010mtld}.  The MLTD roughly reflects a running average of the unique tokens within a body of text, where a higher score represents higher diversity.

As seen in Table~\ref{tab:airline}, Typo Generation outperforms all other techniques in terms of model accuracy, with Paraphrasing and LM Decoding not far behind.  Perhaps unsurprisingly, these three 
also have high number of augmentations generated. On the other hand, Back-translation contains copious duplicates since there are only so ways to translate a sentence while preserving semantics. Curiously, kNN Retrieval also has a low number of augmentations, yet high diversity.  Upon further examination, the kNN model repeatedly retrieves many 
irrelevant items after finding a few high-quality utterances.  This is a direct consequence of using a relatively limited candidate pool.
Among mixed methods, Heuristic Selection exhibits the best classification accuracy. A subtle, but significant consequence of mixing is that all methods have the ability to easily produce the maximum allowed number of augmentations.  Results from the telecommunications domain showcase many of the same patterns, but with LM Decoding edging out Typo Generation as the top accuracy performer (See Table~\ref{tab:telco}).

\paragraph{Human Evaluation} 
Since not all the generated examples are useful or correct, a human evaluation step is still necessary to review the augmentations. The Average Time Spent is the amount of time required to review fifty augmented examples for a single intent when assisted by the data augmentation suggestions.
Acceptance Rate is percent of suggested augmentations that ultimately passed review.   

To study these aspects, we gathered four internal annotators to review the results for all methods on both domains for twelve intents per augmentation technique.  We find that Typo Generation once again performs the best out of all single methods.  However, all the mixed methods have the quick review time and consistently respectable acceptance rates.  Interviewing the annotators reveals that sentence length and complexity, such as from kNN, are the key drivers to increasing time spent. 

\paragraph{Other Considerations} When assessing which augmentation method to deploy, we would ideally consider not just model performance, but also other practical concerns such as ease-of-development and inference latency. 
Translation and LM Decoding take especially long 
since the former requires two passes through the model 
and the latter operates in an auto-regressive manner that is difficult to parallelize. In terms of model complexity, Auxiliary Dataset techniques were noticeably harder to implement due to the dependency on external data sources.  Finally, the complexity of maintaining multiple models may make Mixed Methods not worthwhile to pursue despite their strong performance.

\begin{table}
\centering
\resizebox{\textwidth}{!}{
\begin{tabu}{l|ccccc}
\textbf{Method} &  \textbf{\# Augment} $\uparrow$ & \textbf{Diversity} $\uparrow$ & \textbf{Accuracy} 
        $\uparrow$ & \textbf{Time Spent} $\downarrow$ & \textbf{Accept Rate} $\uparrow$ \\
\tabucline[1.4pt]{-}
Baseline         & ---  & ---  & 26.1 & 14.7 & ---    \\
\hline
EDA              & 5801 & 44.6   & 26.3 &          2.00 & 61.9\%          \\
Synonym          & 5573 & 42.1   & 23.3 &          1.77 & 70.6\%          \\
Paraphrase       & 5876 & 39.5   & 38.8 &          1.42 & 55.6\%          \\
Translation      & 2648 & 41.2   & 28.8 &          1.65 & 65.8\%          \\
Text In-filling  & 5779 & 38.8   & 15.8 &          1.28 & 76.4\%          \\
Typo Generation  & 5824 & 35.2   & 30.4 & \textbf{1.01} & \textbf{77.2\%} \\
kNN Retrieval    & 2446 & \textbf{58.1} &   19.2 & 2.03 & 29.2\%          \\
LM Decoding      & \textbf{5900} & 54.3 & \textbf{45.4} & 1.22 & 63.1\%   \\
\hline
Top 4 Accuracy     &  \textit{5900} &  45.1 &  34.2 & 1.19 & \textit{67.8\%} \\
Category Best      &  \textit{5900} &  45.4 &  30.4 & 1.21 & 60.3\% \\
Heuristic Select   &  \textit{5900} & 39.2 & \textit{40.4} & \textit{1.09} & 60.6\% \\
Mix All            &  \textit{5900} & \textit{46.7} & 29.2 & 1.25 & 67.5\% \\
\end{tabu}}
\caption{Results for telecommunications domain. Fewer intents means max augmentations is 5900.} \label{tab:telco}
\end{table}

\subsection{Analysis and Discussion}
Most methods performed equally well on automatic metrics as on human evaluation. One exception is Text In-filling with relatively low accuracy, but strong acceptance rate.
Digging in, we found in-filling always produces coherent completions, but some of these examples may change the meaning, 
causing the downstream model to learn incorrect associations.  On the other hand, LM Decoding exhibited high accuracy, but low acceptance.
While most intents had high acceptance, certain intents with low coverage in the pre-training stage produced especially poor augmentations, driving down the average.  





\paragraph{Limitations} In general, we found that the size and quality of the seed set to be extremely critical to augmentation success, such that it is quite worthwhile to manually review the seed set before augmenting.  Another limitation of data augmentation comes from the prevalence of duplicated generations, driving down diversity. With that said, optimizing for diversity in isolation can lead to wildly irrelevant augmentations, as evidenced by the output of kNN Retrieval. Finally, the augmentation methods are also limited by their biases, as we did indeed find offensive, racist or otherwise questionable content in a small handful of cases (See Appendix for details).

\section{Conclusion}
We study data augmentation for decreasing time to source and prepare high quality data.
Studying the quantitative results and the trade-offs with qualitative performance and other engineering constraints, we see that certain methods work fairly well on intent classification.
We confirm that in natural language systems, augmentations can be difficult as they may perturb meaning in training data and harm performance.  
Finally, we find that mixing methods are likely to produce strong results, but how to chain together the methods and in what order are untouched avenues left to explore.

\clearpage
 
\begin{ack}
The authors would like to thank Molly Ruhl for her efforts in leading human evaluation, as well as Anna Folinsky and the rest of the data curation team for their participation.
\end{ack}

\bibliography{neurips_2021}

\begin{thebibliography}{21}
\expandafter\ifx\csname natexlab\endcsname\relax\def\natexlab#1{#1}\fi

\bibitem[{Andreas(2020)}]{andreas20good}
Jacob Andreas. 2020.
\newblock \href {https://doi.org/10.18653/v1/2020.acl-main.676} {Good-enough
  compositional data augmentation}.
\newblock In \emph{Proceedings of the 58th Annual Meeting of the Association
  for Computational Linguistics, {ACL} 2020, Online, July 5-10, 2020}, pages
  7556--7566. Association for Computational Linguistics.

\bibitem[{Brown et~al.(2020)Brown, Mann, Ryder, Subbiah, Kaplan, Dhariwal,
  Neelakantan, Shyam, Sastry, Askell, Agarwal, Herbert{-}Voss, Krueger,
  Henighan, Child, Ramesh, Ziegler, Wu, Winter, Hesse, Chen, Sigler, Litwin,
  Gray, Chess, Clark, Berner, McCandlish, Radford, Sutskever, and
  Amodei}]{radford20gpt}
Tom~B. Brown, Benjamin Mann, Nick Ryder, Melanie Subbiah, Jared Kaplan,
  Prafulla Dhariwal, Arvind Neelakantan, Pranav Shyam, Girish Sastry, Amanda
  Askell, Sandhini Agarwal, Ariel Herbert{-}Voss, Gretchen Krueger, Tom
  Henighan, Rewon Child, Aditya Ramesh, Daniel~M. Ziegler, Jeffrey Wu, Clemens
  Winter, Christopher Hesse, Mark Chen, Eric Sigler, Mateusz Litwin, Scott
  Gray, Benjamin Chess, Jack Clark, Christopher Berner, Sam McCandlish, Alec
  Radford, Ilya Sutskever, and Dario Amodei. 2020.
\newblock \href
  {https://proceedings.neurips.cc/paper/2020/hash/1457c0d6bfcb4967418bfb8ac142f64a-Abstract.html}
  {Language models are few-shot learners}.
\newblock In \emph{Advances in Neural Information Processing Systems 33: Annual
  Conference on Neural Information Processing Systems 2020, NeurIPS 2020,
  December 6-12, 2020, virtual}.

\bibitem[{Chen and Yu(2021)}]{chen2021gold}
Derek Chen and Zhou Yu. 2021.
\newblock \href {http://arxiv.org/abs/2109.03079} {{GOLD:} improving
  out-of-scope detection in dialogues using data augmentation}.
\newblock In \emph{Proceedings of the 2021 Conference on Empirical Methods in
  Natural Language Processing, {EMNLP} 2021, Punta Cana, Dominican Republic,
  November 7-11, 2021}, pages 1317--1323. Association for Computational
  Linguistics.

\bibitem[{Devlin et~al.(2018)Devlin, Chang, Lee, and
  Toutanova}]{devlin2018bert}
Jacob Devlin, Ming{-}Wei Chang, Kenton Lee, and Kristina Toutanova. 2018.
\newblock \href {http://arxiv.org/abs/1810.04805} {{BERT:} pre-training of deep
  bidirectional transformers for language understanding}.
\newblock \emph{CoRR}, abs/1810.04805.

\bibitem[{Dolan and Brockett(2005)}]{dolan2005mrpc}
William~B. Dolan and Chris Brockett. 2005.
\newblock \href {https://aclanthology.org/I05-5002/} {Automatically
  constructing a corpus of sentential paraphrases}.
\newblock In \emph{Proceedings of the Third International Workshop on
  Paraphrasing, IWP@IJCNLP 2005, Jeju Island, Korea, October 2005, 2005}. Asian
  Federation of Natural Language Processing.

\bibitem[{Gao et~al.(2020)Gao, Zhang, Ou, and Yu}]{gao20paraphrase}
Silin Gao, Yichi Zhang, Zhijian Ou, and Zhou Yu. 2020.
\newblock \href {https://doi.org/10.18653/v1/2020.acl-main.60} {Paraphrase
  augmented task-oriented dialog generation}.
\newblock In \emph{Proceedings of the 58th Annual Meeting of the Association
  for Computational Linguistics, {ACL} 2020, Online, July 5-10, 2020}, pages
  639--649. Association for Computational Linguistics.

\bibitem[{Hagiwara and Mita(2020)}]{hagiwara2020typo}
Masato Hagiwara and Masato Mita. 2020.
\newblock \href {https://aclanthology.org/2020.lrec-1.835/} {Github typo
  corpus: {A} large-scale multilingual dataset of misspellings and grammatical
  errors}.
\newblock In \emph{Proceedings of The 12th Language Resources and Evaluation
  Conference, {LREC} 2020, Marseille, France, May 11-16, 2020}, pages
  6761--6768. European Language Resources Association.

\bibitem[{Iyer et~al.(2017)Iyer, Dandekar, and Csernai}]{iyer2017qqp}
Shankar Iyer, Nikhil Dandekar, and Kornel Csernai. 2017.
\newblock \href
  {https://www.quora.com/q/quoradata/First-Quora-Dataset-Release-Question-Pairs}
  {First quora dataset release: Question pairs}.
\newblock \emph{Kaggle Competition}.

\bibitem[{Johnson et~al.(2017)Johnson, Douze, and J{\'e}gou}]{johnson2017faiss}
Jeff Johnson, Matthijs Douze, and Herv{\'e} J{\'e}gou. 2017.
\newblock Billion-scale similarity search with gpus.
\newblock \emph{arXiv preprint arXiv:1702.08734}.

\bibitem[{Junczys-Dowmunt et~al.(2018)Junczys-Dowmunt, Grundkiewicz, Dwojak,
  Hoang, Heafield, Neckermann, Seide, Germann, Fikri~Aji, Bogoychev, Martins,
  and Birch}]{junczys2018mariannmt}
Marcin Junczys-Dowmunt, Roman Grundkiewicz, Tomasz Dwojak, Hieu Hoang, Kenneth
  Heafield, Tom Neckermann, Frank Seide, Ulrich Germann, Alham Fikri~Aji,
  Nikolay Bogoychev, Andr\'{e} F.~T. Martins, and Alexandra Birch. 2018.
\newblock \href {http://www.aclweb.org/anthology/P18-4020} {Marian: Fast neural
  machine translation in {C++}}.
\newblock In \emph{Proceedings of ACL 2018, System Demonstrations}, pages
  116--121, Melbourne, Australia. Association for Computational Linguistics.

\bibitem[{Lewis et~al.(2020)Lewis, Liu, Goyal, Ghazvininejad, Mohamed, Levy,
  Stoyanov, and Zettlemoyer}]{lewis2020bart}
Mike Lewis, Yinhan Liu, Naman Goyal, Marjan Ghazvininejad, Abdelrahman Mohamed,
  Omer Levy, Veselin Stoyanov, and Luke Zettlemoyer. 2020.
\newblock \href {https://doi.org/10.18653/v1/2020.acl-main.703} {{BART:}
  denoising sequence-to-sequence pre-training for natural language generation,
  translation, and comprehension}.
\newblock In \emph{Proceedings of the 58th Annual Meeting of the Association
  for Computational Linguistics, {ACL} 2020, Online, July 5-10, 2020}, pages
  7871--7880. Association for Computational Linguistics.

\bibitem[{Liu et~al.(2019)Liu, Ott, Goyal, Du, Joshi, Chen, Levy, Lewis,
  Zettlemoyer, and Stoyanov}]{liu2019roberta}
Yinhan Liu, Myle Ott, Naman Goyal, Jingfei Du, Mandar Joshi, Danqi Chen, Omer
  Levy, Mike Lewis, Luke Zettlemoyer, and Veselin Stoyanov. 2019.
\newblock Roberta: A robustly optimized bert pretraining approach.
\newblock \emph{arXiv preprint arXiv:1907.11692}.

\bibitem[{McCarthy and Jarvis(2010)}]{mccarthy2010mtld}
Philip~M McCarthy and Scott Jarvis. 2010.
\newblock Mtld, vocd-d, and hd-d: A validation study of sophisticated
  approaches to lexical diversity assessment.
\newblock \emph{Behavior research methods}, 42(2):381--392.

\bibitem[{Miller(1995)}]{miller1995wordnet}
George~A. Miller. 1995.
\newblock \href {https://doi.org/10.1145/219717.219748} {Wordnet: A lexical
  database for english}.
\newblock \emph{Commun. ACM}, 38(11):39–41.

\bibitem[{Ng et~al.(2020)Ng, Cho, and Ghassemi}]{ng20ssmba}
Nathan Ng, Kyunghyun Cho, and Marzyeh Ghassemi. 2020.
\newblock \href {https://doi.org/10.18653/v1/2020.emnlp-main.97} {{SSMBA:}
  self-supervised manifold based data augmentation for improving out-of-domain
  robustness}.
\newblock In \emph{Proceedings of the 2020 Conference on Empirical Methods in
  Natural Language Processing, {EMNLP} 2020, Online, November 16-20, 2020},
  pages 1268--1283. Association for Computational Linguistics.

\bibitem[{Niu and Bansal(2019)}]{niu2019auto}
Tong Niu and Mohit Bansal. 2019.
\newblock \href {https://doi.org/10.18653/v1/D19-1132} {Automatically learning
  data augmentation policies for dialogue tasks}.
\newblock In \emph{Proceedings of the 2019 Conference on Empirical Methods in
  Natural Language Processing and the 9th International Joint Conference on
  Natural Language Processing, {EMNLP-IJCNLP} 2019, Hong Kong, China, November
  3-7, 2019}, pages 1317--1323. Association for Computational Linguistics.

\bibitem[{Pennington et~al.(2014)Pennington, Socher, and
  Manning}]{pennington2014glove}
Jeffrey Pennington, Richard Socher, and Christopher~D. Manning. 2014.
\newblock \href {https://doi.org/10.3115/v1/d14-1162} {Glove: Global vectors
  for word representation}.
\newblock In \emph{Proceedings of the 2014 Conference on Empirical Methods in
  Natural Language Processing, {EMNLP} 2014, October 25-29, 2014, Doha, Qatar,
  {A} meeting of SIGDAT, a Special Interest Group of the {ACL}}, pages
  1532--1543. {ACL}.

\bibitem[{Vaswani et~al.(2017)Vaswani, Shazeer, Parmar, Uszkoreit, Jones,
  Gomez, Kaiser, and Polosukhin}]{vaswani2017transformer}
Ashish Vaswani, Noam Shazeer, Niki Parmar, Jakob Uszkoreit, Llion Jones,
  Aidan~N. Gomez, Lukasz Kaiser, and Illia Polosukhin. 2017.
\newblock \href
  {https://proceedings.neurips.cc/paper/2017/hash/3f5ee243547dee91fbd053c1c4a845aa-Abstract.html}
  {Attention is all you need}.
\newblock In \emph{Advances in Neural Information Processing Systems 30: Annual
  Conference on Neural Information Processing Systems 2017, December 4-9, 2017,
  Long Beach, CA, {USA}}, pages 5998--6008.

\bibitem[{Wei and Zou(2019)}]{wei2019eda}
Jason~W. Wei and Kai Zou. 2019.
\newblock \href {https://doi.org/10.18653/v1/D19-1670} {{EDA:} easy data
  augmentation techniques for boosting performance on text classification
  tasks}.
\newblock In \emph{Proceedings of the 2019 Conference on Empirical Methods in
  Natural Language Processing and the 9th International Joint Conference on
  Natural Language Processing, {EMNLP-IJCNLP} 2019, Hong Kong, China, November
  3-7, 2019}, pages 6381--6387. Association for Computational Linguistics.

\bibitem[{Zhang et~al.(2019)Zhang, Baldridge, and He}]{zhang2019paws}
Yuan Zhang, Jason Baldridge, and Luheng He. 2019.
\newblock \href {https://doi.org/10.18653/v1/n19-1131} {{PAWS:} paraphrase
  adversaries from word scrambling}.
\newblock In \emph{Proceedings of the 2019 Conference of the North American
  Chapter of the Association for Computational Linguistics: Human Language
  Technologies, {NAACL-HLT} 2019, Minneapolis, MN, USA, June 2-7, 2019, Volume
  1 (Long and Short Papers)}, pages 1298--1308. Association for Computational
  Linguistics.

\bibitem[{Zhao et~al.(2018)Zhao, Lee, and Esk{\'{e}}nazi}]{zhao18vae}
Tiancheng Zhao, Kyusong Lee, and Maxine Esk{\'{e}}nazi. 2018.
\newblock \href {https://doi.org/10.18653/v1/P18-1101} {Unsupervised discrete
  sentence representation learning for interpretable neural dialog generation}.
\newblock In \emph{Proceedings of the 56th Annual Meeting of the Association
  for Computational Linguistics, {ACL} 2018, Melbourne, Australia, July 15-20,
  2018, Volume 1: Long Papers}, pages 1098--1107. Association for Computational
  Linguistics.

\end{thebibliography}
\bibliographystyle{neurips_natbib}

\newpage
\appendix

\section{Appendix}
\subsection{Qualitative Examples}
As seen in Table~\ref{tab:qual} below, the various methods perform with differing pros and cons.  The intent names have been altered to be more understandable and human readable.   The `Sit Together' intent refers to a user who wants to sit together with someone else on their flight.  The `Change Fees' intent refers to a user inquiring about how much it would cost to update their flight to a different time or location.  Finally, `Opt Out' is when a user wants to opt out of marketing messages from the airline.  

EDA, Synonym and Typo Generation produce un-grammatical outputs with poor syntax, yet can all benefit training.  This implies that robustness to noise is quite important for model training.  Text In-filling will occasionally fill in words that don't make too much sense given the context (i.e. adding ``hungry'' for Opt Out), but still has high acceptance rate.  On the other hand, kNN Retrieval produces grammatical outputs and high diversity, but occasionally harms performance.  This highlights that maintaining the core intent is the only critical task for a data augmentation method. 

\begin{table}[h]
\centering
\resizebox{\textwidth}{!}{
\begin{tabu}{l|ll}
\textbf{Method} &  \textbf{Intent} & \textbf{Augmented Text} \\
\tabucline[1.4pt]{-}
       & Sit Together & way there any is me and my travel party could sit together? \\
 EDA   & Changes Fees & what are the to change a flight? \\
       & Opt Out & can you take off me list mailing your \\
\hline
        & Sit Together & i want to sit with my beau on my flight \\
Synonym & Changes Fees & what cost the fees to change a trajectory ? \\
        & Opt Out & can you involve me off your mailing inclination \\ 
\hline
           & Sit Together & Is it possible to have my wife and I seated next to each other?  \\
Paraphrase & Change Fees & What's the procedure for changing flight tickets? \\
           & Opt Out & I am sick of spam. Just tell me how to stop it. \\
\hline
            & Sit Together & Can you make sure my wife and I are sitting together, please? \\
Translation & Change Fees & Did I hear you got rid of the flight exchange fee? \\
            & Opt Out & I asked you to take off the mailing list a few times, but it didn't \\
            & & \qquad  happen. I'm tired of spam. \\
\hline 
                & Sit Together & i want to be with my boyfriend on my flight \\
Text In-filling & Change Fees & what are the charges if i try to change my flight? \\
                & Opt Out & i have requested multiple times to be taken off then mailing list and \\
                &  & \qquad it has not happened. i am hungry and tired of the spam \\
\hline         
                & Sit Together & Can you make sure me and my wife are satign together, please? \\
Typo Generation & Change Fees & I heard you got rid of flight change reaues? \\
                & Opt Out & I don t not want to get your emails anymore \\
\hline  
              & Sit Together & Can you please link my reservation with my wife and kids? \\
kNN Retrieval & Change Fees & Is it possible to get a refund on the flight? \\
              & Opt Out & I just purchased a ticket and haven’t received the email for it... \\
              & & \qquad think i put down the wrong email. Can you help me? \\
\hline
             & Sit Together & I need to talk to each other about seating arrangements \\
LM Decoding  & Change Fees & Can i change my flight time and pay the difference? \\
             & Opt Out & Sending me an email with my boarding pass \\
\end{tabu}}
\caption{Qualitative examples of augmented text produced by all distinct methods.} \label{tab:qual}
\end{table}

\subsection{Training Setup}
All methods start with the same seed utterances, from which data augmentation proceeds in three steps.  First, the augmentation method is used to generate 3 candidate utterances at a time, to allow for the different methods to cover their own hyper-parameters.  For example, EDA allows for insertions, deletions, or swaps.  So one of each augmentation type is generated in the candidate set.  Second, we pass the candidates to a diversity ranker which calculates the BLEU score of the set of utterances if we were to add the candidate.  The candidate which results in the lowest BLEU score (and thus highest diversity) is kept for consideration.  In the final step, the candidate is compared as an exact match against the seed data and previously added augmentations.  If the candidate is unique, then it is added to the final pool of augmentations.  If the candidate is a repeat of a previous augmentation, then we retry the augmentation process.  If 10 retries are accumulated for a given data augmentation method, the generation process terminates.  This explains why certain methods (e.g. kNN) contains much fewer augmentation examples than others.

Each method trained as a fine-tuned RoBERTa-base classifier.  The models are trained for up to 14 epochs with early stopping if there was no improvement for 5 consecutive epochs.  The hyperparameters we tune include learning rate, dropout rate and occasionally temperature.   The batch size was kept constant at 16.  We found learning rates between 1e-5 and 3e-4 to work well across methods.  Dropout rate was tested among [0.0, 0.05, 0.1].  Each method received the same amount of tuning (6 rounds) to ensure fairness across methods.  Each round of training took roughly 15-20 minutes on a Nvidia Tesla-V100 GPU, which was used for all experiments.  This was accessed through Amazon as AWS EC2 instances.


\end{document}